\documentclass[10pt,twocolumn,letterpaper]{article}

\usepackage{iccv}
\usepackage{times}
\usepackage{epsfig}
\usepackage{graphicx}
\usepackage{amsmath}
\usepackage{amssymb}
\usepackage{authblk}
\usepackage[breaklinks=true,bookmarks=false]{hyperref}

\iccvfinalcopy % *** Uncomment this line for the final submission

\def\iccvPaperID{12707} % *** Enter the ICCV Paper ID here

\ificcvfinal\fi

\begin{document}

%%%%%%%%% TITLE
\title{Confidence-aware 3D Gaze Estimation and Evaluation Metric}

\author{Qiaojie Zheng }
\author{Xiaoli Zhang}
\affil{Colorado School of Mines, Golden CO 80401, USA}
\affil{\tt\small zheng@mines.edu,xlzhang@mines.edu}

% {\tt\small zheng@mines.edu \and \tt\small zhangjiucai@gmail.com \and \tt\small 2143886@jeffcoschools.us \and \tt\small xlzhang@mines.edu}
% \and

\maketitle
\ificcvfinal\thispagestyle{empty}\fi

%%%%%%%%% ABSTRACT
\begin{abstract}
   Deep learning appearance-based 3D gaze estimation is gaining popularity due to its minimal hardware requirements and being free of constraint. Unreliable and overconfident inferences, however, still limit the adoption of this gaze estimation method. To address the unreliable and overconfident issues, we introduce a confidence-aware model that predicts uncertainties together with gaze angle estimations. We also introduce a novel effectiveness evaluation method based on the causality between eye feature degradation and the rise in inference uncertainty to assess the uncertainty estimation. Our confidence-aware model demonstrates reliable uncertainty estimations while providing angular estimation accuracies on par with the state-of-the-art. Compared with the existing statistical uncertainty-angular-error evaluation metric, the proposed effectiveness evaluation approach can more effectively judge inferred uncertainties' performance at each prediction.  
\end{abstract}

%%%%%%%%% BODY TEXT
\section{Introduction}

The simplicity in hardware requirements and constraint-free settings make appearance-based gaze estimation attractive for human-machine-interaction (HMI) applications, such as virtual reality \cite{Clay2019}, driver monitoring systems \cite{Palazzi2019}, and assistive robotic arms \cite{Sunny2021, Cio2019}. Typical fine-angle appearance-based 3D gaze estimation comprises two stages. The first stage adopts facial landmark detection techniques to crop out eye image patches and extract head angles. The second stage uses the cropped eye image patches and head angles to infer pitch and yaw gaze angles measured from the center of the eyes. Recent advancements in deep learning (DL), especially convolutional neural networks (CNN), significantly improve gaze angle inference accuracy and robustness. The state-of-the-art methods\cite{Fischer2018,Park2019} achieve average angular accuracies of around 3-8 degrees in occlusion-free and constraint-free datasets, such as MPII \cite{Zhang2015} and RTGene \cite{Fischer2018}. 
%%%%

\begin{figure}[t]
\begin{center}
\includegraphics[width=0.95\linewidth]{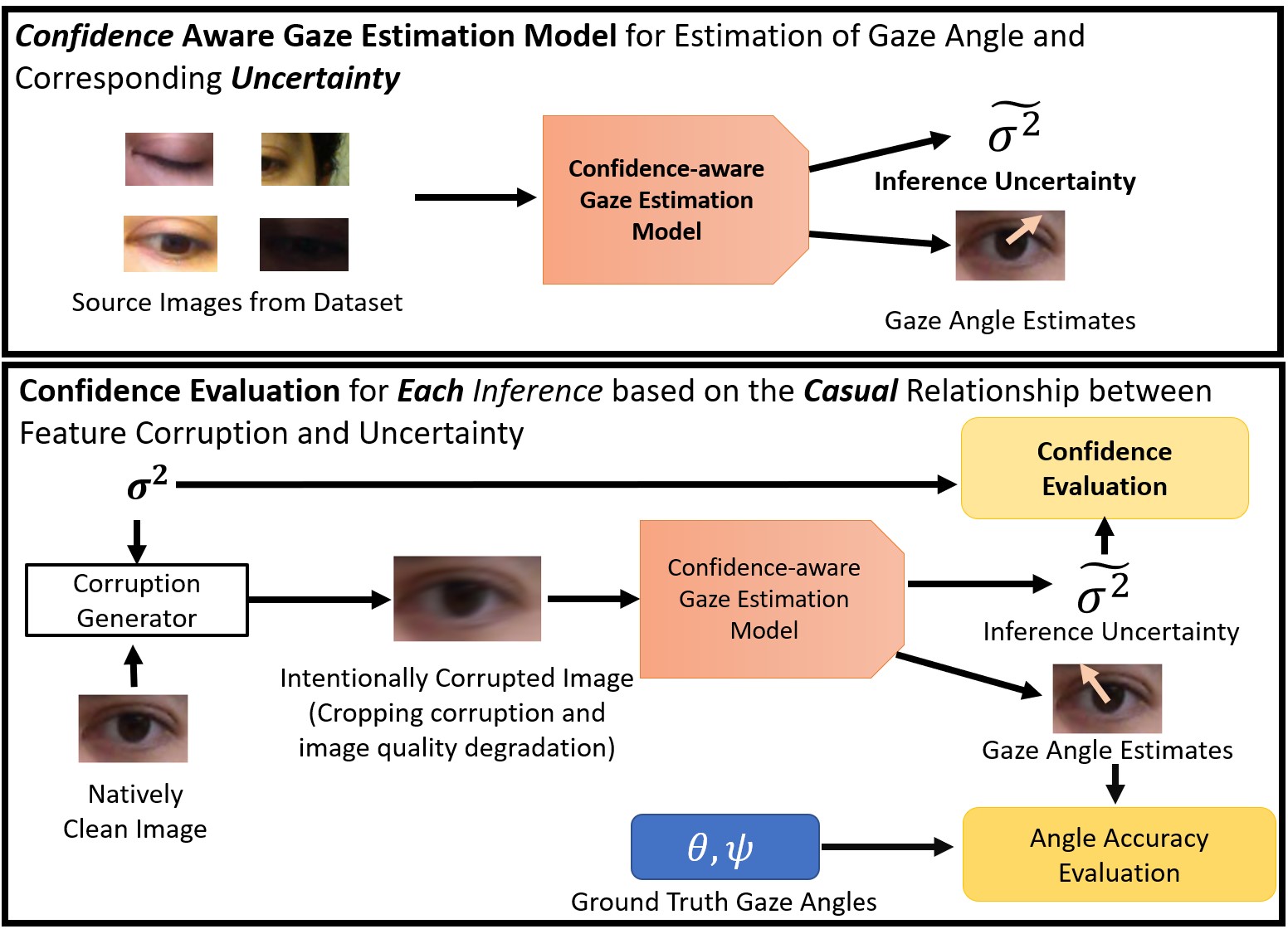}
\end{center}
   \caption{The proposed confidence-aware model (top) and the uncertainty effectiveness evaluation approach (bottom). Our model learns to judge the prediction confidence based on eye feature quality in the input images with our proposed loss function. Inference uncertainties are produced together with gaze angle estimates. Our uncertainty effectiveness assessment is based on the asserted causality between eye feature degradation and inference uncertainty. We assess the effectiveness based on the correlation strength between the inferred uncertainty and the severity of intentionally introduced corruptions used to achieve different levels of eye feature degradation. }
\label{fig:Figure_1}
\end{figure}

However, challenges still need to be addressed in handling inaccurate predictions due to the large variability in image quality and individual appearance differences. Existing methods mainly focus on improving angular estimation accuracies but overlook the prediction uncertainties caused by these challenges. These overlooked uncertainty effects will cause catastrophic problems and limit their adoption. For example, in situations where the eye features in images have been heavily corrupted or eliminated, the DL methods will still output unreliable and erroneous gaze estimation and be highly confident about it. Subsequent HMI applications that use these inaccurate estimations may behave unpredictably, losing human trust. To avoid unforeseeable overconfident inference, most HMI applications that critically depend on appearance-based gaze estimation are still performed under controlled environments \cite{Sunny2021, Li2019} Further adoptions of appearance-based gaze estimations require confidence output in addition to the angular estimation value to inform subsequent decision-making processes about potential errors.

In addition, a competent evaluation approach is lacking to assess the effectiveness of the estimated uncertainty values. Existing evaluation methods only focus on the overall performance of statistical angular estimation errors and ignore and are not capable of evaluating the effectiveness of individual uncertainty inference. Moreover, the uncertainty-angular-error correlation used by existing methods is non-causal. In other words, high inference uncertainty does not necessarily lead to large errors in predicted gaze angles. The inference-error-confidence correlation may be meaningful to evaluate the performance at a statistical level but not feasible to assess the confidence at each individual inference.

To fill these gaps, this paper proposes a confidence-aware model and new procedures with novel metrics to evaluate the effectiveness of the estimated uncertainties as shown in Figure \ref{fig:Figure_1}. The confidence-ware model addresses the overconfident inference problem by outputting numerical values for prediction uncertainties together with the original gaze estimates. The proposed model learns detrimental influence factors, such as closed eyes, that corrupt input images and assigns high uncertainty values for their gaze angle estimates. A specially designed loss function enables unsupervised learning for these detrimental features without ground truth labelling. This paper also takes a step further and proposes a novel, more effective evaluation method and metrics to assess the effectiveness of confidence awareness. The proposed evaluation approach is based on evaluating the causal relationship between the severities of intentionally introduced corruptions and the models; inferred uncertainties. This evaluation approach addresses the limitations of the non-causal uncertainty-angular-error correlation used by the existing method. In short, the contributions of this work can be summarized as follows:

\begin{enumerate}
  \item A confidence-aware model that outputs numerical values for inference uncertainties is proposed. This model can achieve angular accuracies on par with the state-of-the-art method while giving confidence estimates of the inferred gaze angle for subsequent applications.
  \item A novel and more effective evaluation approach and metrics to assess the effectiveness of the uncertainty output from the proposed model. The proposed approach introduces a causal metric that measures the correlation between the severities of intentionally introduced corruptions and the model's inferred uncertainty value.
  \item Extensive experiments are conducted to demonstrate the advantage of uncertainty estimation and the effectiveness of the uncertainty estimation method. A qualitative evaluation verifies the causality assumption between eye feature degradation and inferred uncertainty. 
\end{enumerate}

%-------------------------------------------------------------------------

\section{Related Work}
% This section overviews existing methods in deep learning appearance-based 3D gaze estimations, general confidence estimation approaches, confidence estimates in gaze-related applications, and evaluation approaches for assessing the effectiveness of inferred confidence.

\subsection{Deep learning 3D Gaze Estimations}
Zhang et al. \cite{Zhang2015} first introduced deep convolutional neural network model together with their constraint-free MPII-Gaze dataset. This network takes eye-region image patches to perform gaze vector inference. In their later work \cite{Zhang2017}, Zhang et al. proposed a new spatial weight network design that takes the full-face image for gaze angle inference. Kellnhofer et al. \cite{Kellnhofer2019} proposed the Gaze360 dataset, which samples gaze angles in all 360 degrees in outdoor conditions. They also introduced a long-short-term-memory network that takes seven consecutive frames for gaze analysis. Fischer et al. \cite{Fischer2018} proposed the RTGene dataset that captures subjects at much greater distances than the MPII dataset. Their proposed gaze estimation network used the VGG16 network for feature extractions and implemented an ensemble inference scheme for added robustness and accuracy. Yu et al. \cite{Yu2020} proposed the first unsupervised gaze representation learning structure and showed a strong linear correlation between the learned gaze representation and the ground truth angles. Other network designs and learning approaches, such as dilated convolution \cite{Chen2019} and meta-learning \cite{Park2019}, are also proposed. Most network structure or method improvements in these works aimed to lower the angular prediction error; not much attention was devoted to estimating uncertainties related to the predictions  

\subsection{Confidence Estimation Approaches}
Recent studies have proposed many uncertainty quantifications approaches in deep learning, including Bayesian Neural Networks (BNNs) \cite{Gal2015}, heteroskedastic maximum likelihood estimation \cite{Nix1994}, etc. A combination of the Bayesian approach and heteroskedastic MLE can be used to distinguish epistemic and aleatoric uncertainties in some situations \cite{Kendall2017}. The Bayesian approach, including its dropout approximations, places probability on the network's weight, thus concluding the epistemic uncertainty. BNN inferencing requires multiple passes, which slows down the prediction process. The heteroskedastic MLE approach relates uncertainty to the input data, thus concluding the aleatoric uncertainty. Unlike BNN inferencing, networks trained with heteroskedastic MLE only require a single pass for inferencing, thus preserving the run time of a regular neural network.

\subsection{Confidence Estimation in Gaze Tracking}
There are very few works that incorporated confidence estimation into gaze estimation tasks. All these methods are applied to coarse gaze estimations based on the head or facial information rather than detailed eye features used in this paper. Their uncertainty sources majorly come from eye region occlusion rather than various image corruptions studied by this work. In \cite{Dias2020}, they perform gaze estimation based on low dimensional facial landmarks and the confidence estimated by the preceding OpenPose\cite{Cao2017} anatomical key points detecting method. They proposed a Confidence Gated Unit to incorporate confidence information from OpenPose for gaze estimations. The output dimension was expanded from 2 to 3 to accommodate uncertainty estimations. A heteroscedastic MLE with a cosine-similarity-based negative log-likelihood loss function was used to train the model. In \cite{Kellnhofer2019}, the authors applied quantile regression, test time dropout, and test time augmentation to various network structures with CNN backbone to estimate confidence ranges for gaze prediction.

\subsection{Evaluation of Confidence Effectiveness}
Due to the lack of ground truth labels for confidence, there is no simple numerical value to be directly compared to evaluate the performance of confidence estimation in gaze estimation tasks. Previously mentioned methods in \cite{Kellnhofer2019} and \cite{Dias2020} used the correlation between the estimated uncertainties and prediction angular errors. This correlation, however, may not be trustworthy because it is non-causal. These two works' inference confidence changes are caused mainly by eye occlusion rather than inference angular error. Thus, a causal effectiveness measure would relate occlusion severity to inference uncertainty. Although the current uncertainty-angular-error correlation showed a positive correlation between average angular errors and the estimated uncertainties, this value can only be used to statistically demonstrate that there are more samples with high angular estimation errors in the high-uncertainty group. Such positive correlations do not necessarily hold true for individual sample tests because individual samples can have high uncertainty and low angular error at the same time. In other words, a perfect uncertainty model would not achieve a perfect correlation of 1 in this evaluation. A method based on causal correlation is needed to measure the confidence-aware feature more effectively.
%-------------------------------------------------------------------------

\section{Methodology}
\subsection{Confidence-aware gaze estimation network structure and training}
\subsubsection{Confidence-Aware Network Structure}
\begin{figure}[t]
\begin{center}
\includegraphics[width=0.9\linewidth]{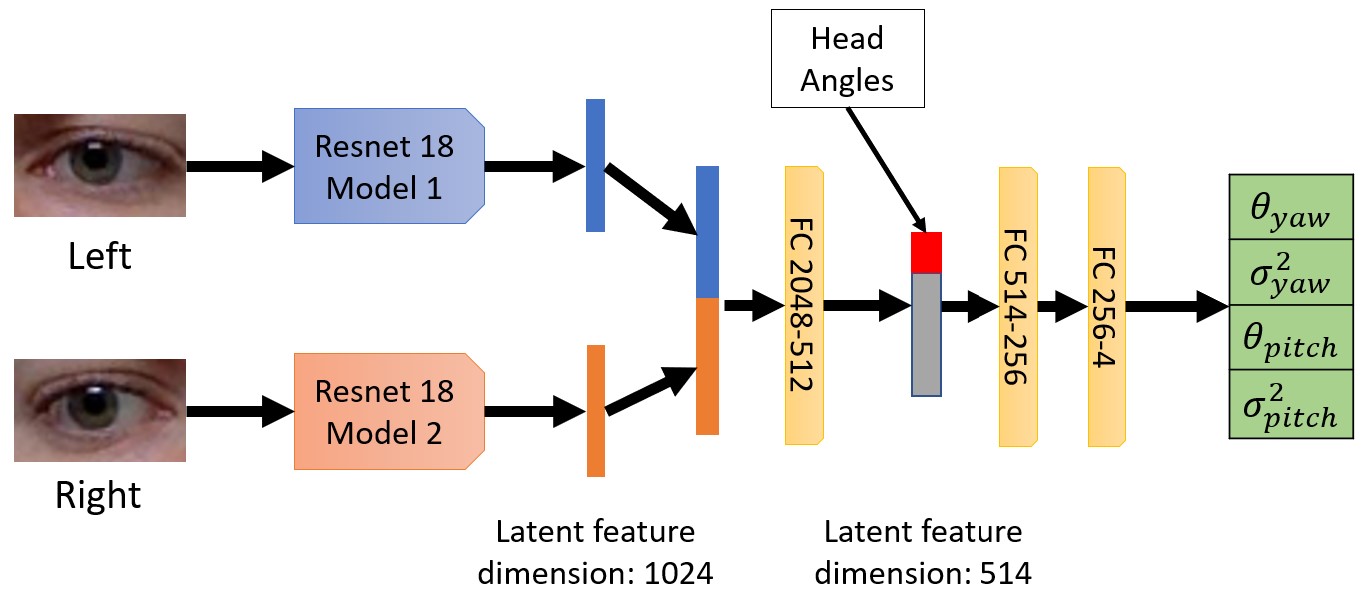}
\end{center}
   \caption{Network structure adapted from \cite{Fischer2018} for confidence-aware 3D gaze estimation. This network outputs uncertainty values for pitch and yaw angles, respectively. The maximum between the pitch uncertainty and yaw uncertainty represents the overall inference uncertainty. }
\label{fig:Figure_2}
\end{figure}

Recall those appearance-based gaze estimations included two stages: eye region detection and angular gaze estimation with cropped eye image patches. Since the facial landmark detection used by eye region extraction, such as \cite{Baltrusaitis2014}, is a relatively mature technology, this paper will not present architectures used by facial landmark detection. It will mainly focus on the confidence-aware gaze angle estimation part. The model takes cropped eye patches and head angle information and adopts deep neural networks to predict both gaze angles (pitch and yaw) and their uncertainties, as shown in Figure \ref{fig:Figure_2}.

The confidence awareness comes from the heteroskedastic assumption of the input data. Each set of inputs into the network is asserted to have unique associated variances to the outputs, which are treated as inference uncertainties. To accommodate the heteroskedastic assumption in 3D gaze angle estimation, the proposed network outputs 4 values, 2 for yaw and pitch angle estimation and the other 2 for their associated inference uncertainties. The maximum between pitch and yaw uncertainties represents the overall inference uncertainty. 

The proposed network structure is adapted from the work of Fischer et al. Left and right eye images are first fed into Resnet18 models for feature extraction. We chose Resnet18 for its small size and fast training time while maintaining comparable output accuracies. The extracted features are represented by a 1024 vector for each eye. The extracted features are then concatenated and passed through a series of fully connected (FC) layers to perform inference. The head angle vector, which is a 1x2 vector containing head pitch and yaw angles, is concatenated with the output from the first FC layer to be considered for inferencing. 

\subsubsection{Loss Function}
The proposed model is fully differentiable and can be trained end-to-end with only labels for gaze angles. We minimize a customized compounded loss (Equation \ref{eq:NIX_original}) modified from \cite{Nix1994} containing two parts; an uncertainty-regulated angular error loss term $\frac{l_n}{2 \sigma^2\left(x\right)}$ for gaze inference accuracies and an uncertainty regularization term $\frac{1}{2} \ln \left(\sigma^2\left(x\right)\right)$ to avoid unbounded uncertainty regulation to the angular error loss. 
\begin{equation}
  \operatorname{loss}=\frac{1}{2} \ln \left(\sigma^2\left(x\right)\right)+\frac{l_n}{2 \sigma^2\left(x\right)}
  \label{eq:NIX_original}
\end{equation}
The uncertainty-regulated angular loss contains an angular loss term $l_n$ in the numerator to represent the difference in inference and ground truth value and an uncertainty term  $\sigma^2(\cdot)$  in the denominator to calculate uncertainty based on input data $x$. When the model outputs angle prediction with large errors, the uncertainty will be increased to lower the overall loss value. To avoid the infinite growth of uncertainty, a regularization term on its natural log values is needed in the overall loss function. To avoid gradient explosion and achieve training stability, the angular loss is calculated from smooth L1 loss depicted in equation \ref{eq:smooth_l1}. 
\begin{equation}
  l_{n,t}=\left\{\begin{array}{cc}
0.5(H_\theta(x_t)-\theta_t)^2, & \text { for }|H_\theta(x_t)-\theta|<1 \\
|H_\theta(x_t)-\theta_t|-0.5, & \text { otherwise }
\end{array}\right.
  \label{eq:smooth_l1}
\end{equation}

\subsection{Evaluation of Confidence Awareness}
\begin{figure}[t]
\begin{center}
\includegraphics[width=0.9\linewidth]{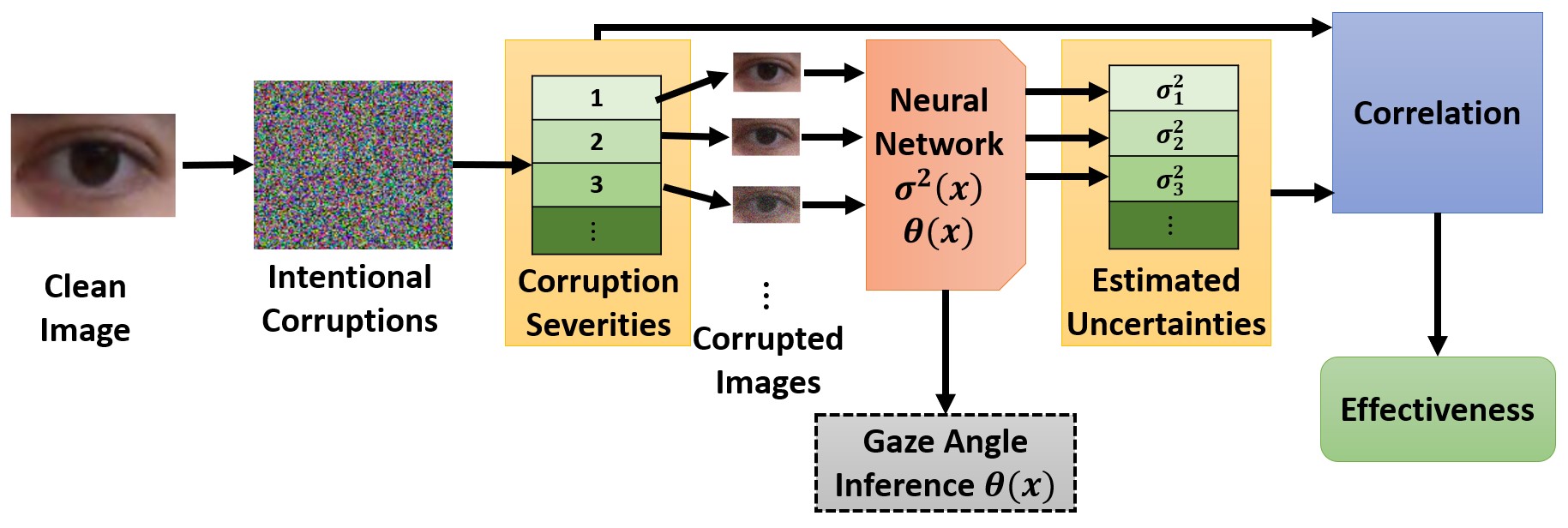}
\end{center}
   \caption{Proposed procedure to evaluate effectiveness in uncertainty estimation. Intentional corruption with controllable severities is introduced to clean images for the confidence-aware model to infer. The model's estimated uncertainties are compared with the corruption severity levels to find effectiveness. }
\label{fig:Figure_3}
\end{figure}

To measure the effectiveness of the network's confidence awareness efficiently and accurately, we propose a novel evaluation approach that depends on a causal relationship between image feature degradations and uncertainties, which we refer to as corruptions and inferred uncertainties. We introduce controllable corruptions with different levels of severities to relatively clean images with little to no corruptions and pass these intentionally corrupted images into the confidence-aware network. We evaluate confidence awareness based on the correlation between inferred uncertainties and the corruption level (Figure \ref{fig:Figure_3}).

\subsubsection{Image Feature Degradation Definition}
In the gaze estimation application, the image feature degradation caused the input images to contain unfamiliar features for the model to infer. Two types of degradation on images are used in gaze estimation: improper image handling and source-level degradation. 

Improper image handling attributes the cause of degradation to the process of acquiring eye patch images, which contain general image degradation, such as blurring and noises, and gaze-tracking-specific degradation, such as eye region off-cropping. During image handling, these degradations are applied to a clean source. Clearer eye region images could be captured if improper handling was avoided.

Source level degradation attributes the cause of degradation to the subject from whom the images are collected. Typical source-level degradation can be closed eyes or drastic eye shape differences. The source-level degradation cannot be further reduced due to the corrupted source.   
\subsubsection{Two Assumptions}
The proposed evaluation method is based on two assumptions: 1) most of the training samples are relatively clean from corruption, and 2) inferred uncertainties are positively correlated  to the severity of corruption.

The first assumption lets the model learn the image-to-gaze-angle mapping function based on clear eye features. If the training dataset contains too many heavily corrupted images, the model cannot learn about the functionality of eye features in gaze estimation. The inferred uncertainties, therefore, cannot reflect confidence in gaze angle estimations. The second assumption enables us to quantitatively evaluate the model's uncertainty estimation performance. A perfectly trained confidence-aware model should output uncertainty values with a strict positive correlation with the introduced severity level. 

\subsubsection{Proposed Method and Metric}
\label{method and metric}
Based on the two assumptions, we propose the evaluation method as follows:
\begin{enumerate}
  \item From a dataset, choose images with low inferred uncertainties by the model and visually check the image to ensure it contains clear eye features. The visual check will avoid datasets that have been uniformly corrupted in every image. 
  \item Apply predefined controllable corruptions at different severities to the clean image and pass these images to the model for uncertainty inference.
  \item Calculate the correlation between the severity of introduced corruptions and inferred uncertainty. The correlation should be close to 1 for uncertainty estimations with good performance.
\end{enumerate}

With this evaluation method, we also propose an evaluation metric (Equation \ref{eq:metric}) to calculate the overall performance based on all types of corruptions introduced to the image. In Equation \ref{eq:metric}, $C_i$ is Spearman's rank correlation coefficient for the $i^{th}$ introduced corruption. $k_i$ is the slope between the severity of introduced corruptions and inferred uncertainty when fitted with linear regression. A small slope magnitude represents that the model is robust against specific corruption, thus, does not show much fluctuation in the inferred uncertainty. We would like to make the correlation contribution from the low-impact corruption small by scaling its correlation value by its slope. $n$ denotes the total types of corruption introduced. 

\begin{equation}
  P=\frac{\sum_i^n k_i C_i}{\sum_i^n\left|k_i\right|}
  \label{eq:metric}
\end{equation}

% \begin{figure}[t]
% \begin{center}
% \includegraphics[width=0.6\linewidth]{Figures/Figure_4_new.jpg}
% \end{center}
%    \caption{Effectiveness metric illustration with a hypothetical corruption. X axis indicates corruption severity applied to images, and larger values indicate more eye feature degradation. Y axis indicates the inferred uncertainty, and larger values mean uncertainty prediction. The 6 data points show that the inferred uncertainty positively correlates with the corruption, and therefore, have a Spearman’s rank coefficient of 1. The slope is used to weigh contributions of correlation when multiple types of corruptions are studied.  }
% \label{fig:Figure_4}
% \end{figure}

\section{Experiment Setup}
We build our model with PyTorch and evaluate the proposed confidence-aware model in 3D gaze estimation tasks and its uncertainty estimation performance with two open-source datasets, the MPII-Gaze and RTGene datasets. The experiment setup and experiment results are described in the following paragraphs.
% \subsubsection{Datasets}
% MPII and RTGene are the two datasets used in this study. The MPII dataset was used for training the confidence-aware model due to its higher image quality, and the RTGene dataset was used for cross-dataset tests. The MPII dataset contains 213,659 images collected from 15 participants, and the RTGene dataset contains 122,531 labelled images collected from 15 participants recorded in 17 sessions. These two datasets contain left and right eye image patches, pitch and yaw angles for head movement, and pitch and yaw angles for gaze direction as ground truth labels. These two datasets were selected because of their similarities in data collection settings – indoors without head movement constraints. Their dissimilarity in image quality could also be meaningful for testing confidence awareness. 

% Because the MPII dataset was recorded at a shorter distance with only minimal post-processing, such as eye cropping, it contains eye images with higher quality compared to those from the RTGene dataset. On the other hand, the RTGene dataset was recorded at a greater distance with complex pos-processing, such as eye tracking glass removal through Generative Adversarial Network (GAN); it contains images of low quality with heavy corruptions. The image quality difference essentially puts the images collected into two different domains regarding noise corruption while the recorded ground truth angles are measured similarly.

\subsection{Training Settings and Hyperparameters}
All model training is performed on the MPII dataset. The initial learning rate was set to 0.0001 with a weight decay factor of 0.1 after epoch 25. Adam optimizer and batch size of 64 was used. Network weights are initialized to the pre-trained ones from ImageNet. 
A leave-one-out training-testing split strategy was applied for the within-dataset test. Images from 14 out of the 15 participants were used for training and validation. We followed an 80-20 splitting rule to distribute the training and validation data. The input image patch was prepared following the method in RTGene by first resizing eye patch images from $36\times60$ to $224\times224$ to use the Resnet18 structure better. The color channels are then normalized with means of 0.485, 0.456, and 0.406 with standard deviations of 0.229, 0.224, and 0.225, respectively. No data augmentation, such as random crop or color jitter, was performed.

\subsection{Corruption Methods in Effectiveness Evaluations}
Recall that corruptions can be categorized into two groups based on the cause – corruptions due to improper image handling and corruptions due to bad sources. Since it would be nearly impossible to quantify the corruption extent that natively exists in datasets, we designed experiments to simulate their effects of feature degradation with controllable corruptions whose severities are known. 

\begin{figure}[t]
\begin{center}
\includegraphics[width=0.78\linewidth]{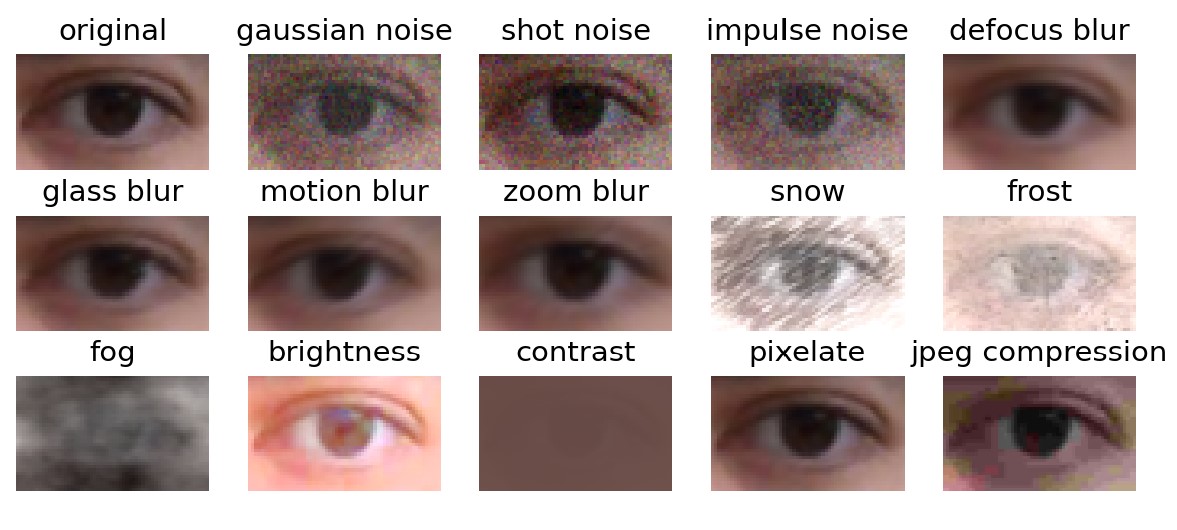}
\end{center}
   \caption{Visualization of 14 image corruptions. The top left shows the uncorrupted image for reference. These corruptions are adopted from ImageNet-C \cite{Hendrycks2019} Elastic transform corruption is not used because of its inconsistent behavior with the corruption severity.  }
\label{fig:Figure_5}
\end{figure}

The general image quality degradation and source-level corruptions are simulated with 14 out of the 15 corruption methods proposed in ImageNet-C\cite{Hendrycks2019} The elastic transform is excluded because it does not output visually consistent degradation with the severity levels. These 14 corruptions capture various degradations to feature sharpness, such as the eyelid and iris boundary, which compose most of the effect caused by general image quality degradation and source-level corruption. These 14 corruptions are simulated with the 5 severity levels as described in ImageNet-C.  Figure \ref{fig:Figure_5} shows all 14 corruptions with the highest corruption severity and the uncorrupted image for comparison.

The gaze-tracking application-specific corruptions typically leave images with sharp eye features but with a partial or entire cutoff of the overall features. These corruptions cannot be reflected by the previously mentioned 14 and require custom implementation. Therefore, a custom implementation of vertical and horizontal eye patch off-cropping is developed to simulate the application-specific off-cropping corruption. These off-cropping corruptions are achieved by intentionally moving the cropping window away from the eye center by predefined distances. Five levels of severities are designed for this type of corruption to cover slight off-cropping to total off-cropping of the eye regions. The off-cropping effects are shown in Figure 5.

\begin{figure}[t]
\begin{center}
\includegraphics[width=0.9\linewidth]{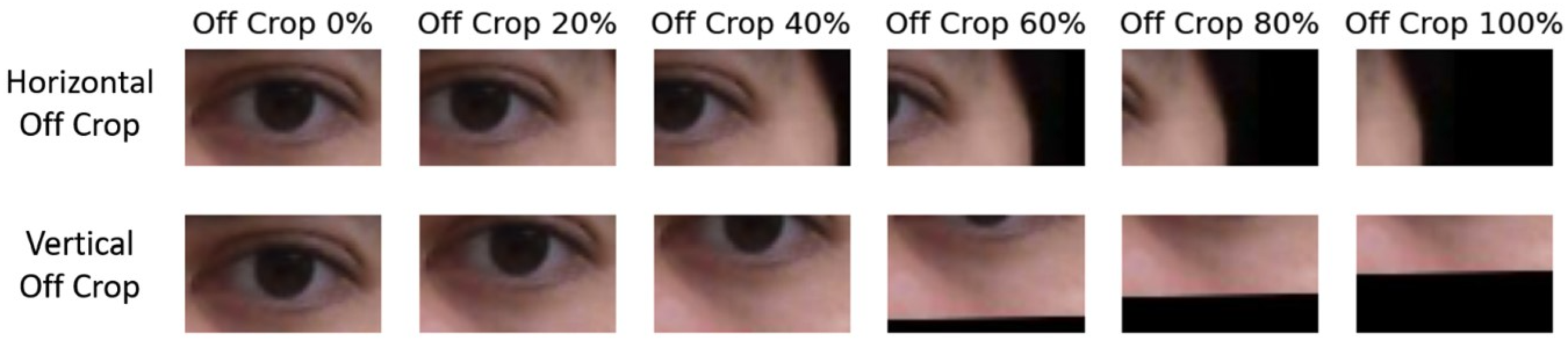}
\end{center}
   \caption{Custom implementation of off-cropping image corruption with 5 levels of severity. The leftmost column contains uncorrupted images. The most severe off-cropping completely crops eye features by moving the crop center by the width or height of the patch. The rest 4 severities are spaced evenly by the crop center distance between no off-crop and the most severe one.  }
\label{fig:Figure_6}
\end{figure}

\subsection{Evaluation Metrics}
The qualitative evaluation metric will be based on equation \ref{eq:metric} In this experiment setup, we consider all introduced corruptions. That is, $n$ in equation \ref{eq:metric} is 16. When the model can effectively capture inference uncertainty, the correlation score calculated from equation \ref{eq:metric} will be close to 1. Exist effectiveness evaluation method, which calculates the correlation between angular inference error and inferred uncertainty, will also be calculated as baseline values to be compared.

Lastly, a qualitative evaluation is present in determining the effectiveness of the confidence-aware algorithm in evaluating corruptions that natively occurred in the dataset. Images are sorted by the model's inferred uncertainty value. Selected images from each confidence quantile are displayed on plots for human judgment. The qualitative evaluation is also used to verify the causality assumption between the corruption severity and inferred uncertainty magnitude.   

\section{Result}
\subsection{Design of Experiments}
The experiments are designed to study 1) the effectiveness of confidence awareness of the newly proposed confidence aware model on different corruptions, 2) the differences between the new and existing evaluating approach, 3) the generalizability of the model and evaluation metric across person and dataset, and 4) qualitatively judge the confidence-aware model for unquantifiable corruptions that occur natively in the dataset. 
\subsection{Experiment Result}
\subsubsection{Measuring Effectiveness of Confidence-Aware Model on Each Corruption}
\begin{figure}[t]
\begin{center}
\includegraphics[width=1.0\linewidth]{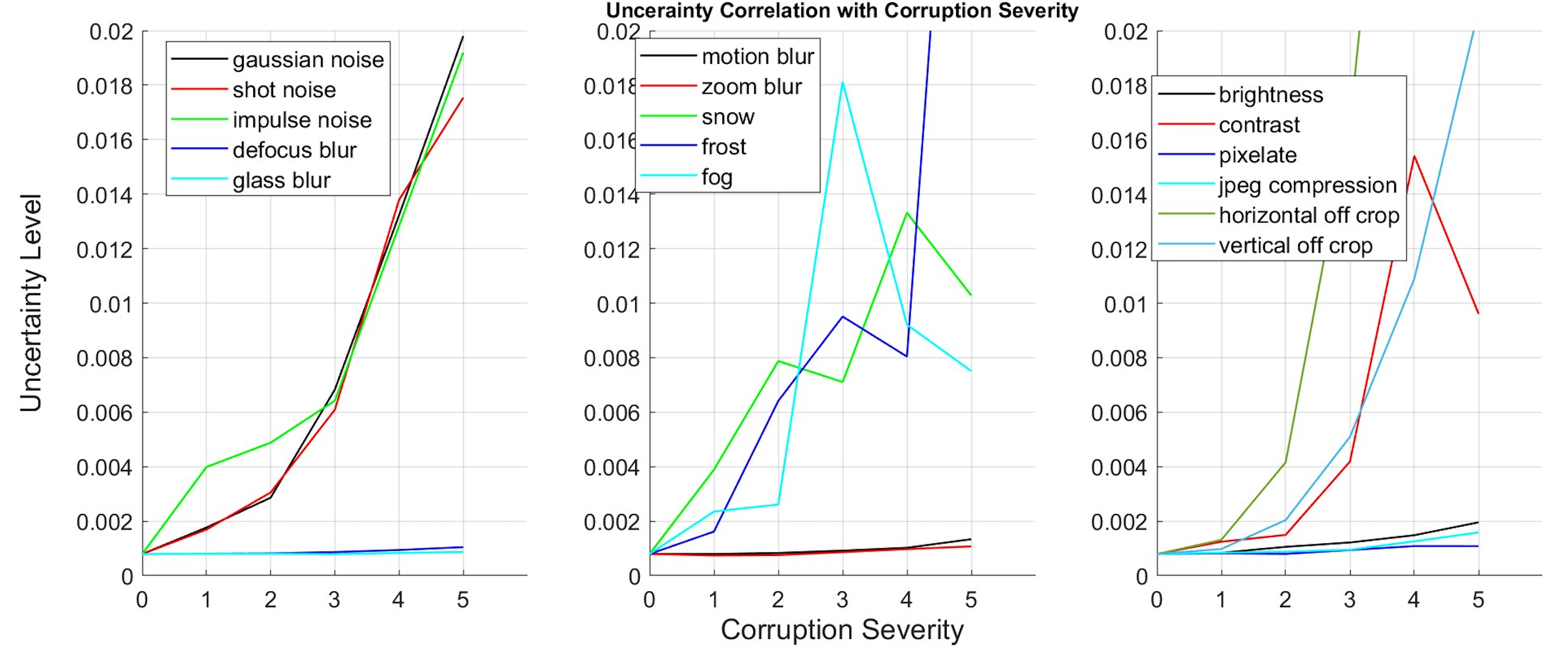}
\end{center}
   \caption{Correlation behaviors between the corruption severities and the inferred uncertainties on 16 intentionally introduced corruptions In MPII dataset. The introduced corruptions can be grouped into two categories based on the model's behavior - insensitive corruptions and sensitive corruptions. Model's behaviors on insensitive corruptions are shown as lines with small slopes and consistently low uncertainty values. Light feature degradation in these corruptions caused such behaviors. Model's behaviors on sensitive corruptions are depicted as large slopes with some inconsistency when corruption severity is beyond 3. Heavy feature degradation caused the large slope. The inconsistent trends in medium to high severities from the sensitive corruptions are caused by eye feature overwhelming. Correlation score calculated using equation \ref{eq:metric} is 0.9534, suggesting very strong correlation. }
\label{fig:Figure_7}
\end{figure}

The effectiveness of the confidence-aware model on single corruptions is judged following the same concept using the correlation score and slope described in \ref{method and metric}. We first analyze the effectiveness by plotting the corruption severity against the inferred uncertainty to visualize the correlation magnitude and slope in Figure \ref{fig:Figure_7}. Datapoints in this figure were obtained by performing inference on a single image that is relatively clean (image 13322, person 0 in MPII dataset). Because we assume that uncertainties are proportional to the severities of introduced corruptions, we expect a perfect model to output uncertainties that are strictly positively correlated with corruption severity. 
Among these trendlines for corruption, several show near-zero slopes because the model is relatively robust and insensitive to these corruptions. These corruptions include defocus blur, glass blur, motion blur, zoom blur, brightness change, pixelation, and JPEG compression. Figure \ref{fig:Figure_5} shows that these corruptions cause very light eye feature degradation compared to some of the heavy ones. These degradations are not severe enough to cause high uncertainties for model inferencing. 

All other trendlines show that the model is sensitive to the rest of the corruptions. All these corruptions caused heavy eye feature degradation or even elimination in some extreme situations. These corruptions caused the model to be unable to extract desired eye features and output high uncertainties. Some trendlines displayed a downward trend near the end, such as snow, because the eye features are overwhelmed by corruptions causing the model to perform inference outside of the designed region. Since very few samples in the training dataset contain such features, the model does not have enough knowledge to estimate the corresponding uncertainties correctly. It should be noted that although the uncertainty values dipped at these locations, their magnitude is still well above the baseline values, indicating non-trustworthy gaze angle estimation. 

Using the newly designed evaluation metric in equation \ref{eq:metric}, the performance on effectiveness is 0.9534 ($>0.8$), demonstrating strong capability in detecting unreliable inferences. This high correlation effectiveness score is on par with trendline behaviors unaffected by the model's robustness against some corruptions.
\subsubsection{Comparison with Baseline Evaluation Method }
\begin{figure}[t]
\begin{center}
\includegraphics[width=1.0\linewidth]{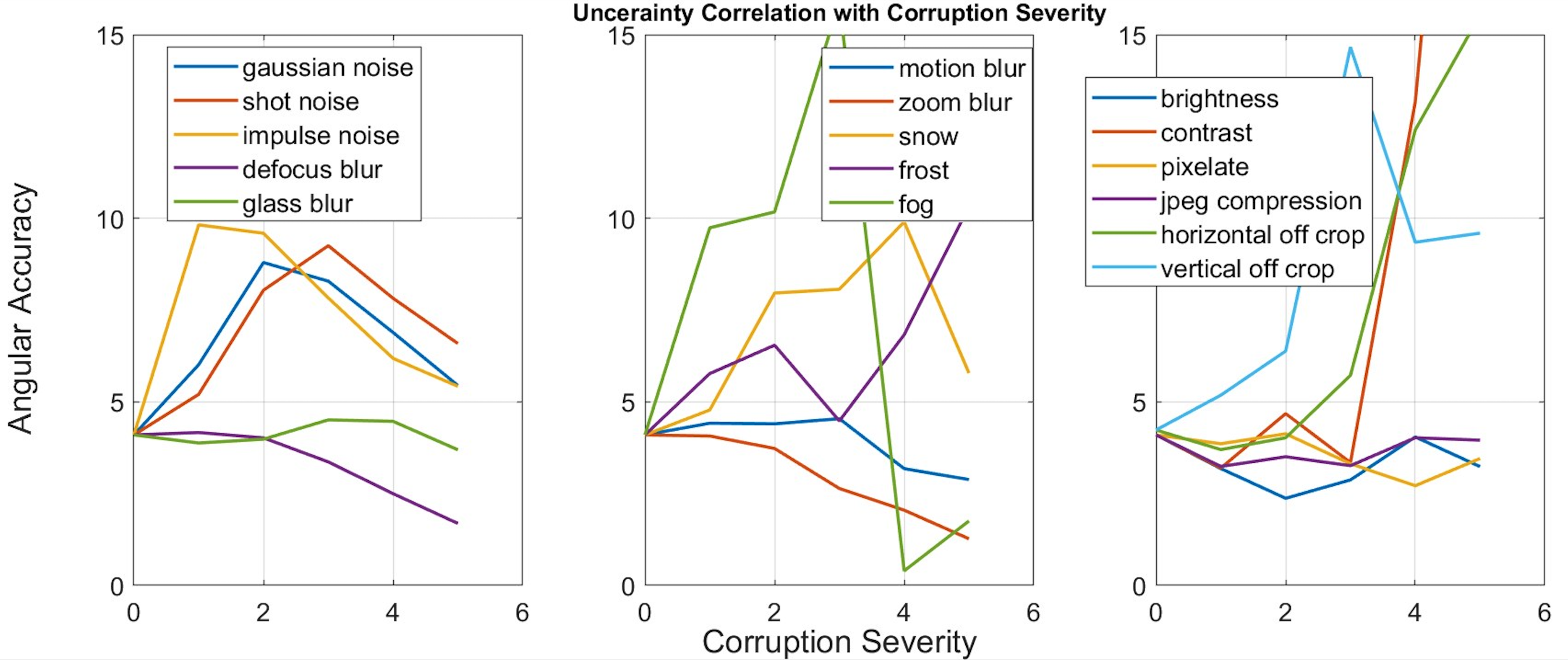}
\end{center}
   \caption{Correlation behaviors between corruption severity and prediction angular accuracy on 16 intentionally introduced corruptions. The trendline are much less consistent than those from Figure \ref{fig:Figure_7} where correlations between corruption severity and inferred uncertainties are studied. The lack of consistency suggests a weak correlation between angular accuracies and the severity of introduced corruption. A lower correlation score of 0.7109 calculated from equation \ref{eq:metric} also suggests a lower correlation. }
\label{fig:Figure_8}
\end{figure}

To compare the effectiveness between the existing and new proposed evaluation methods for uncertainty estimates, we plot the angular error values against introduced corruption severity to judge the correlation strength with the uncertainty source in Figure \ref{fig:Figure_8}. This figure is intended to study the correlation strength between angular errors and the root causes for inference uncertainties. Although general upward trends can be observed for all uncertainty values, trendline behaviors for the "sensitive corruptions" are different and much less consistent across the board. Using the proposed correlation calculation in equation \ref{eq:metric}, we calculated the correlation strength between the angular errors and severity levels to be 0.7109. The less consistent behaviors in Figure \ref{fig:Figure_8} and the lower correlation number suggest that angular prediction errors correlate with uncertainty sources weakly. Since the inferred uncertainty should be caused by image corruption severity, this low correlation score suggests that correlating angular values to inferred uncertainty may not reflect the model's actual performance. 

\begin{figure}[t]
\begin{center}
\includegraphics[width=0.95\linewidth]{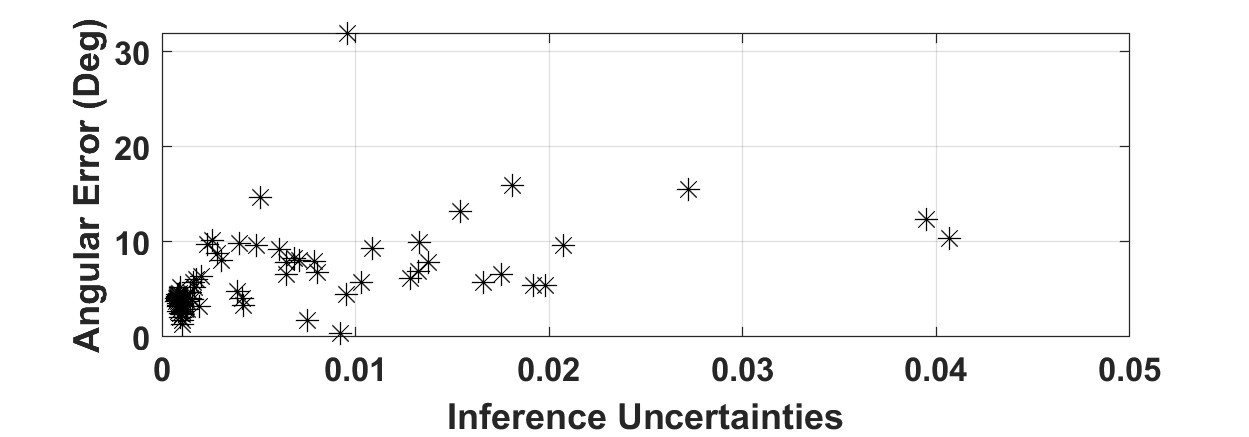}
\end{center}
   \caption{Scatter plot for visualizing the correlation between prediction uncertainties and the prediction angular error. Data points in this figure do not show a strong trend. The Spearman's rank correlation score was calculated to be 0.5384, suggesting a medium to low correlation between these two values.   }
\label{fig:Figure_9}
\end{figure}

In addition to calculating the correlation strength using the proposed method, we also performed a correlation study that links inferred uncertainty to angular errors as used in \cite{Kellnhofer2019, Dias2020}. The overall correlation is displayed in a scatter plot in Figure \ref{fig:Figure_9}. No apparent trend can be discovered between the uncertainty and angular error. The Spearman's correlation calculated with the data points shown in Figure \ref{fig:Figure_9} is 0.5384, which suggests a medium to low correlation. Therefore, using angular errors to evaluate the effectiveness of a model is not representative. 

\subsubsection{Consistency Study}
\begin{figure}[t]
\begin{center}
\includegraphics[width=0.9\linewidth]{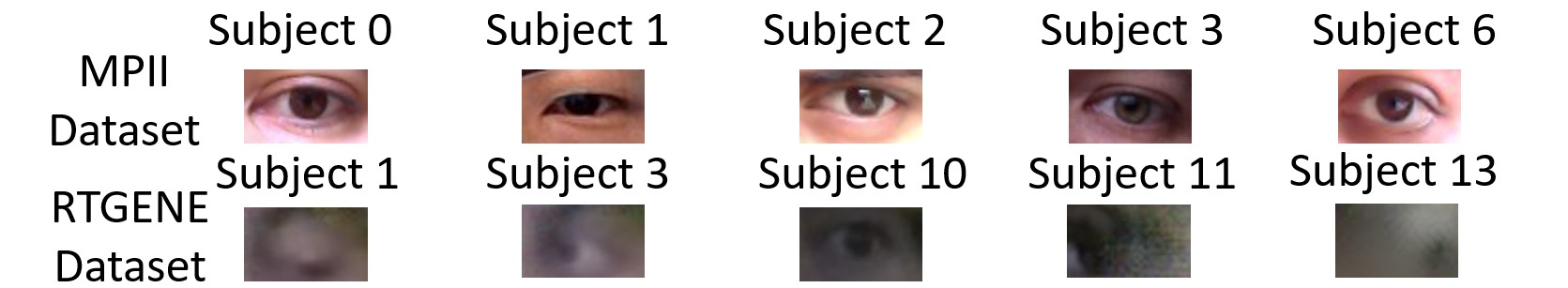}
\end{center}
   \caption{Comparison of the test subjects from two datasets}
\label{fig:Figure_10}
\end{figure}

Confidence awareness must be able to achieve consistent performance across different people to be applicable. This consistency study used a leave-one-out strategy to study within-dataset variations and used weights trained on MPII to perform cross-dataset study on RTGene. Due to the great number of test combinations, this paper only selects two corruptions, motion blur and contrast degradation, to represent the light hand heavy feature degradation corruption among all types. These two corruptions are applied to 100 images randomly selected from the upper 20\% confidence percentile, i.e., images with 20\% lowest uncertainties, to ensure the original images are relatively free of preexisting noises. These images are selected from 5 testing subjects with the most image data and are shown in Figure \ref{fig:Figure_10}.

\begin{figure}[t]
\begin{center}
\includegraphics[width=0.85\linewidth]{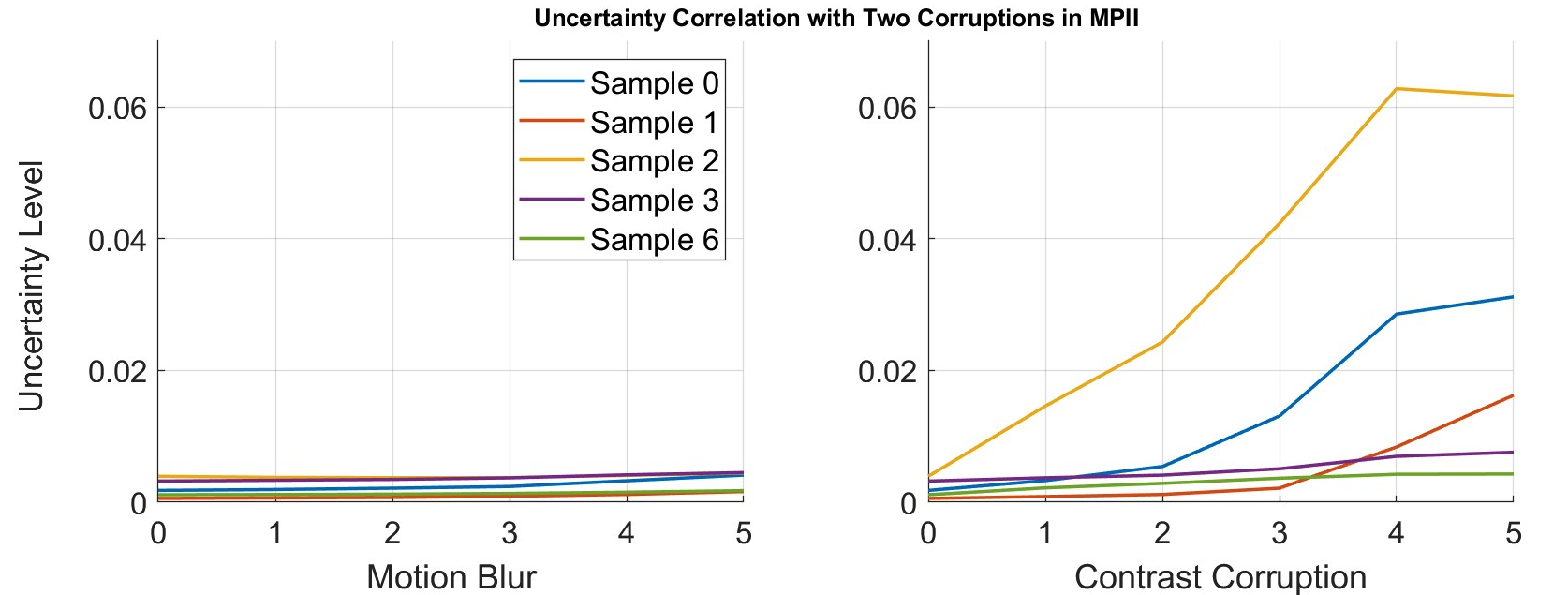}
\end{center}
   \caption{Within dataset consistency for confidence awareness in the MPII dataset. 5 samples with the most images are studies with light (motion blur) and heavy (contrast) corruptions. The model shows consistent behavior for light corruption due to the eye feature preserving motion blur corruption. The model shows different behaviors when images were processed with heavy corruption due to the overwhelming of the eye features by the corruption. The overwhelm of corruption feature caused the model to infer outside the training range, leading to inconsistent results. Trends in both corruptions resemble those from Figure \ref{fig:Figure_7} where single image is used. This similarity suggests that the model can achieve consistent behaviors across samples within dataset. }
\label{fig:Figure_11}
\end{figure}

\begin{figure}[t]
\begin{center}
\includegraphics[width=0.85\linewidth]{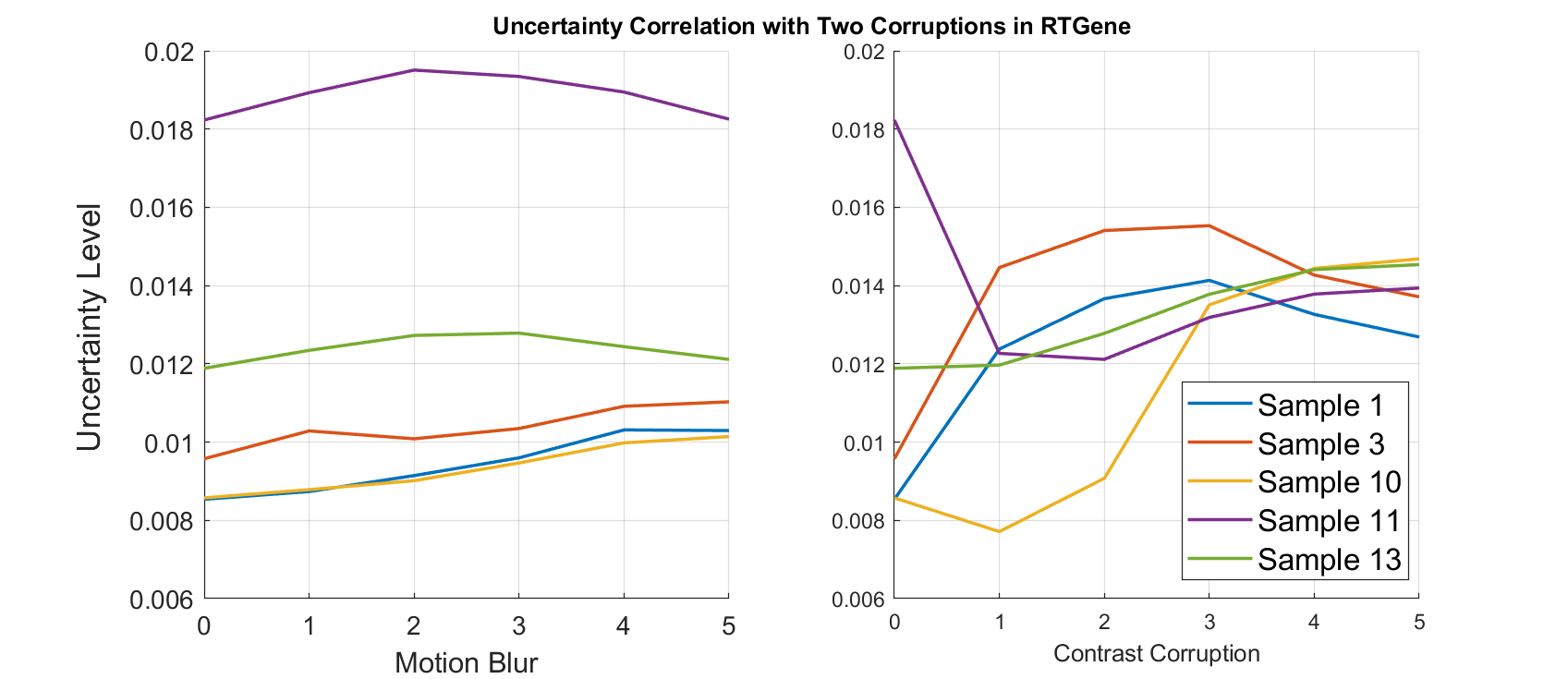}
\end{center}
   \caption{Correlation behaviors between the corruption severities and the inferred uncertainties from the RTGene dataset. The model is trained with MPII dataset. 5 test subjects with the most images are used for testing with light and heavy corruption. The trendline behavior differs significantly from that of MPII due to the high level of preexisting corruptions. High values in the predicted uncertainty level suggest that the gaze angle prediction result using models trained on MPII is very unreliable on RTGene }
\label{fig:Figure_12}
\end{figure}

First, results from the cross-person consistency study are presented in Figure \ref{fig:Figure_11}. The trendline for light corruption (motion blur on the left) shows highly similar behavior across all samples. This similarity is due to the feature-preserving nature of the motion blur corruption. On the other hand, trendlines diverge for heavy corruption at high severity. The diverge is caused by out-of-range inference since training datasets rarely have completely corrupted images without eye features. Similarities in trends between Figure \ref{fig:Figure_7}, where single image is used, and Figure \ref{fig:Figure_12} suggest the consistency across samples within dataset. The effectiveness scores for each of the 5 test subjects using all 16 corruptions are calculated from equation \ref{eq:metric}, shown in Table \ref{tab:table_1}.

 All effectiveness scores are relatively high, indicating an effective confidence-aware model. It should be noted that the effectiveness score for test subject 1 is relatively lower than the rest. This is caused by the difference in eye features caused by eye shape difference. As shown in the top row of Figure \ref{fig:Figure_10}, eye features from sample 1 are not as obvious as the rest and may cause the confidence-aware model to be less effective on test sample 1. Due to the scarcity of similar eye features in the training dataset, the model performs inference with higher uncertainties on subject 1.

\begin{table}
\begin{center}
\begin{tabular}{|l|c|c|c|c|c|}
\hline
Subject     & 0      & 1      & 2      & 3      & 6      \\ \hline
Correlation & 0.95 & 0.85 & 0.91 & 0.91 & 0.94 \\ \hline

\end{tabular}
\end{center}
\caption{Uncertainty Effectiveness Score for subjects from MPII}
\label{tab:table_1}
\end{table}

Next, the results of cross-dataset consistency study are present in Figure \ref{fig:Figure_12}. The model's behavior on the RTGene dataset  shows different behaviors than that of MPII dataset. Due to the high corruption levels natively exists in all samples in the RTGene dataset, as shown in the bottom row of Figure \ref{fig:Figure_10}, the model's estimated uncertainties in the uncorrupted are much higher than that from RTGene. The preexisting corruptions also caused the model's behavior to be less consistent than MPII. The high uncertainty values suggest that model inferencing result performed on RTGene should not be trustworthy. 

\begin{figure}[t]
\begin{center}
\includegraphics[width=0.85\linewidth]{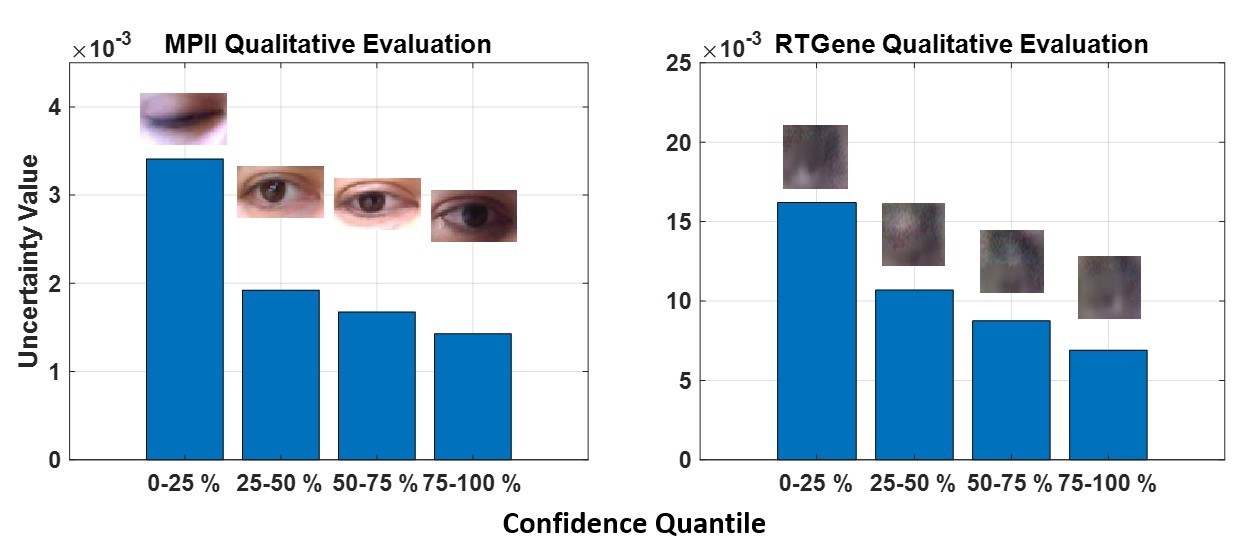}
\end{center}
   \caption{Qualitative evaluation of the model on MPII and RTGene datasets with weights trained on the MPII dataset. Based on left plot, the model can successfully distinguish heavily corrupted images from the rest. In the right plot, although all images are affected by very heavy corruption, the model can still put images with slight eye features in the most confident quantile. The overall uncertainty in the right figure is almost 10 times as large as the left one, suggesting not trustworthy inference results.  }
\label{fig:Figure_13}
\end{figure}
\subsubsection{Quanlitative Evaluation}
To verify the effectiveness of the confidence-aware model against the preexisting unquantifiable corruption, we performed qualitative evaluations to examine the corruptions that exist in each confidence quantile visually. Figure \ref{fig:Figure_13} shows a qualitative evaluation by displaying selected images from each confidence quantile. The lowest confidence quantile contains images with the most severe corruptions and has much higher quantile-wise average values compared with the later three quantiles. Common corruptions in the lowest confidence quantile involve closed eyes, complete off-cropping, or drastic lightning condition change. A cross-dataset qualitative evaluation is conducted on RTGene dataset with weights trained on MPII Because images in the RTGene datasets are corrupted with much severity noises, the overall uncertainty magnitude is about 10 times that from MPII. The most confident quantile in the right of Figure \ref{fig:Figure_13} still shows certain eye features, albeit heavily corrupted.

It should be noted that the most confident quantile in the RTGene evaluation has higher uncertainty scores than the least confident quantile in the MPII, which contains little to no eye features. Based on this comparison, any inferences performed on RTGene dataset with the MPII-trained weights should not be trustworthy. This quantitative evaluation result also verified that our assumption on the causation between corruption severity and inference confidence is correct in the proposed confidence-aware model. Higher uncertainty values are assigned to the images with heavy corruption even with unquantifiable corruption from other test participants or dataset.  
\section{Conclusion}
This work introduced an effective confidence-aware gaze estimation model against image corruption and a novel, accurate evaluation approach for determining the effectiveness of confidence awareness on each inference. The evaluation approach involves intentionally introducing controllable corruptions, whose severities correlate with the inference confidence for effectiveness evaluation. The model shows consistent performance behavior across samples and dataset. The confidence-aware model has demonstrated its capability with the newly proposed evaluation methods. This confidence-aware model can make HMI safer by avoiding passing erroneous gaze information to the machine and improving the adoption rate for critical applications.  

{\small
\bibliographystyle{ieee_fullname}
\bibliography{gaze_related}
}

\end{document}